\documentclass[preprint, review, 10pt, times, twocolumn, 5p]{elsarticle}

\usepackage{longtable}
\usepackage{float}
\usepackage{adjustbox}
\usepackage{xltabular}
\usepackage{booktabs}
\usepackage{caption}
\usepackage{url}
\usepackage{graphicx}
\usepackage[colorlinks=true, allcolors=blue]{hyperref}
\usepackage{setspace}

\begin{document}
\singlespacing

\begin{abstract}

Digital Twins have recently gained attention in various industries for simulation, monitoring and decision-making purposes because most of them rely on ever-improving machine learning models in their architecture. However, agricultural Digital Twin implementations are still limited compared to other industries. Meanwhile, machine learning in general, and reinforcement learning in particular, have demonstrated their potential in agricultural applications like optimising decision-making processes, task automatisation and resource management.\\
A key aspect of Digital Twins is the representation of physical assets or systems in a virtual environment. This characteristic synergises well with the requirements for reinforcement learning, which relies on environment representations to accurately learn the best policy for a given task. Therefore, the use of reinforcement learning in agriculture has the potential to open up a variety of reinforcement learning-based Digital Twin applications in agricultural domains.\\
To explore these domains, this review aims to categorise existing research works that employ reinforcement learning techniques in agricultural settings. On the one hand, categories are created regarding the application domain, such as robotics, greenhouse management, irrigation systems, and crop management, identifying the potential future application areas for reinforcement learning-based Digital Twins. On the other hand, the reinforcement learning techniques employed in these applications, including tabular methods, Deep Q-Networks (DQN), Policy Gradient methods, and Actor-Critic algorithms, are categorised to gain an overview of currently employed models.\\
Through this analysis, the review seeks to provide insights into the current state-of-the-art in integrating Digital Twins and reinforcement learning in agriculture. Additionally, it aims to identify gaps and opportunities for future research, including potential synergies of reinforcement learning and Digital Twins to tackle agricultural challenges and optimise farming processes, paving the way for more efficient and sustainable farming methodologies.
\end{abstract}


\begin{frontmatter}


\title{Current applications and potential future directions of reinforcement learning-based Digital Twins in agriculture}



\author[2,3]{Georg Goldenits\corref{cor1}}

\cortext[cor1]{Corresponding author}
\ead{ggoldenits@sba-research.org}

\author[3]{Kevin Mallinger}

\author[1]{Sebastian Raubitzek}

\author[1,3]{Thomas Neubauer}

\affiliation[1]{organization = {SBA Research gGmbH}, 
                addressline = {Floragasse 7/5.OG},
                city = {Vienna},
                postcode = {1040},
                state = {Vienna},
                country = {Austria}}

\affiliation[2]{organization = {University of Vienna}, 
                addressline = {Universitätsring 1},
                city = {Vienna},
                postcode = {1010},
                state = {Vienna},
                country = {Austria}}

\affiliation[3]{organization = {TU Wien}, 
                addressline = {Karlsplatz 13},
                city = {Vienna},
                postcode = {1040},
                state = {Vienna},
                country = {Austria}}




\end{frontmatter}



\section{Introduction}
\noindent
Growing concerns about food security driven by population growth and increasing climate variability have raised the pressure for more productive and efficient farming \cite{Intro1}\cite{Intro2}\cite{Intro3}\cite{Intro4}. A recent research approach to optimising farming operations has been the introduction of Digital Twins in agricultural applications. Digital Twins replicate a real entity in a virtual representation and allow for simulating and optimising tasks and events supported by machine learning models \cite{DTDef1}. Simulation is especially useful for scenarios where change would otherwise only be observed over a long period of time, such as crop growth, or where the risk of taking incorrect actions is high, for example, in incorrect irrigation management that could lead to losses in crop yield. This practice can also support sustainability in agriculture and help maintain or increase crop yields. Even though there already exist Digital Twins for automated harvesting using unmanned vehicles and irrigation management tasks, potential other areas of application remain unexplored \cite{DTAgRev2}\cite{DTAgRev6}.\\
As current Digital Twins rely on machine learning, a potential way to identify additional use cases for them is to look at tasks already solved by machine learning, but so far, Digital Twins have not been implemented. One machine learning technique that lends itself well to simulation is reinforcement learning. It can be used model-free and self-learn how to handle a situation based on predefined parameter and environment settings. These properties also allow reinforcement learning to adapt to unseen situations while trying to achieve the initially defined goal \cite{RLDef}. Therefore, there are already existing reinforcement learning implementations in agriculture, such as suggesting using resources like water and fertiliser more efficiently \cite{water1}\cite{water2}, increasing crop yields by detecting pests and diseases using unmanned aerial vehicles \cite{croma1}\cite{robo1}, planting crops in a suitable order \cite{croma8} or reduce the energy consumption of greenhouses \cite{green1}\cite{green2}.\\
A significant factor for the applicability of results achieved by a reinforcement learning agent is how well the environment it interacts with has been modelled after the real environment \cite{croma12}. Therefore, research on Digital Twins and how to best replicate the real world aligns with the needs of a well-trained reinforcement learning agent and could pave the way for reinforcement learning-based Digital Twins in agriculture. The term reinforcement learning-based Digital Twin is used in this review to define Digital Twins that rely on reinforcement learning as their machine learning model.\\
To identify promising reinforcement learning-based Digital Twin applications in agriculture, this review aims to categorise recent applications of reinforcement learning in agriculture and seeks to provide a structured overview of them. For each category, the strengths and weaknesses of reinforcement learning compared to other possible solutions are discussed, and an assessment is made if and how it could be implemented in a Digital Twin. The obtained factual insights can inform future research directions and contribute to developing and implementing advanced agricultural management systems.
The goals are summarised in the following research questions:
\begin{itemize}
    \item RQ1 What are the existing applications of reinforcement learning in agriculture?
    
    \item RQ2 Which application domains are suitable for reinforcement learning-based Digital Twins?
\end{itemize}
The work will be structured as follows:
Section 2 will introduce Digital Twins and reinforcement learning by defining them. Section 3 describes the methodology of this work. Section 4 summarises related literature reviews on Digital Twins in agriculture and machine learning in agriculture. Section 5 attempts to categorise the application areas for reinforcement learning in agriculture. The potential for reinforcement learning-based Digital Twins is assessed based on the strengths and weaknesses of reinforcement learning for applications in each category. Also, an outlook on potential future applications is given. Section 6 provides the conclusion of the manuscript.

\section{Definitions}
\noindent
Michael Grieves first introduced Digital Twins in 2003\cite{DTDef} with the goal of optimising a factory process. The general idea of Digital Twins is to replicate a real-world object, entity or system, such as a train, a crop or an agricultural supply chain in a virtual environment. Sensor data is commonly used in an Internet of Things (IoT) setting to create a digital image because it allows continuous measuring of a system or specific properties. The data gets processed, and frequently, a machine learning model is used to draw inferences from the collected data and use it for predicting actions and conditions of the environment.\\
Once a Digital Twin is in place, simulations are commonly used to explore how an environment would behave in different circumstances. This allows the user or an automated system to be better prepared for various scenarios and react appropriately in the real world should a situation arise that was previously only simulated. Of course, it may be difficult to replicate real-world applications based on the task perfectly. Therefore, model outputs and simulation results might deviate from the corresponding real-world situation. However, in a functioning Digital Twin system, real-world experiences get fed back into the Digital Twin, leading to updates in models and simulations \cite{DTDef1}.\\
\newline
According to Sutton \& Barto\cite{RLDef}, reinforcement learning is learning what to do. Within the machine learning world, reinforcement learning is a concept that does not fit in the classical categories of supervised or unsupervised learning but instead represents its own learning category.\\
In general, reinforcement learning tries to optimise a sequence of actions that may be previously unknown within a given environment by collecting rewards it obtains by interacting with the environment. The goal is to maximise the reward signal and, in doing so, arrive at an optimal course of action. In the classical approach to reinforcement learning, a table for each state-action pair is maintained, and the quality of each pair is incrementally updated using the Bellman Equation\cite{RLDef1}. More recent developments combine reinforcement learning with artificial neural networks (ANN) to approximate the quality of a state-action pair by minimising the error between the predicted value and the target value. Since the ANNs used to support reinforcement learning may vary in structure and complexity, those models are summarised as deep reinforcement learning (DRL).\\
Compared to tabular reinforcement learning, DRL can handle much larger state spaces as the function approximation is computationally more efficient, and no table of all possible combinations of states and actions needs to be maintained. While DRL, in most cases, approximated the optimal solution well, tabular reinforcement learning achieves the optimal solution given enough time and, due to its maintained table, allows for easier explainability of its decision-making process.

\section{Methodology}
\begin{figure}[ht]
\begin{center}
\includegraphics[width=0.5\textwidth]{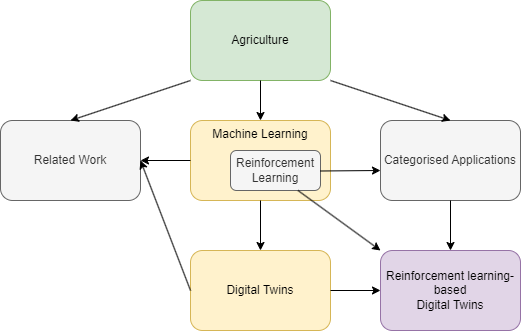}
\caption{\label{fig:RS_Scope}Research scope of this work}

\end{center}
\end{figure}

\noindent
This review aims to categorise current research in agricultural reinforcement learning by area of application and type of model used. For each manually defined category, strengths and weaknesses are highlighted. Only works published in 2020 or later are considered to account for current research.\\
This review will summarise the areas of Digital Twins in agriculture and general machine learning in agriculture in the related work section to gain a broader picture of the current state of research. The scope of this work can be summarised in Figure~\hyperref[fig:RS_Scope]{\ref{fig:RS_Scope}}. The search queries can be found in Table~\hyperref[tab:Queries]{\ref{tab:Queries}}. The queries are used to search for published work online on Google Scholar (https://scholar.google.com/). As of the last access date (Feb 7th 2024), searching for publications yielded more than 17,000 results. Therefore, only the first 200 publications listed are used as a sample for the categorisation as mentioned earlier, resulting in a corpus of 71 publications.\\
\newline
\newline

\begin{table}[!ht]
\caption{\label{tab:Queries}}
\begin{tabular}{@{}l@{}}
\toprule
\multicolumn{1}{c}{\textbf{Search queries}}                                                               \\ \midrule
\begin{tabular}[c]{@{}l@{}}"Digital Twin" AND "Agriculture": Related Work \end{tabular} \\ \midrule
\begin{tabular}[c]{@{}l@{}} "Machine Learning" AND "Agriculture": Related Work\end{tabular} \\ \midrule
\begin{tabular}[c]{@{}l@{}}"Reinforcement Learning" AND "Agriculture": Main Part\end{tabular} \\ \bottomrule
\end{tabular}
\end{table}

\noindent
The exclusion criteria for publications are:
\begin{itemize}
    \item Only papers where the full text is accessible and written in English are considered.
    
    \item Review papers, aside from those in the related work section, are excluded
    
    \item As this work aims to capture the state-of-the-art, only papers published from 2020 onward are considered.

    \item Publications that do not contribute novel ideas to reinforcement learning but instead only cite it as related work in similar research areas are excluded

    \item Only peer-reviewed publications are considered
\end{itemize}

\section{Related Work}
\noindent
To gain a broad overview of the current state-of-the-art for Digital Twins in agriculture, existing areas of application and current challenges identified in related review literature are summarised. With a similar goal in mind, machine learning solutions for problems in agriculture are determined based on existing review literature in that area. Aside from gaining an overview, the summary provides a foundation to assess the strengths and weaknesses of reinforcement learning implementations compared to other machine learning solutions.

\subsection{Digital Twins in agriculture}
\noindent
Digital Twins in agriculture are becoming a popular topic as the potential of representing crops, automated robots or entire farming systems virtually with the goal of optimising processes is realised. The increasing popularity is exemplified by seven review papers that were published in 2022 and 2023 and can be found in Table~\hyperref[tab:RW_1]{\ref{tab:RW_1}}. These papers summarise recent advances and present current challenges and opportunities.

\begin{table}[!h]
\caption{\label{tab:RW_1} Corpus of papers reviewed in Section 4.1}
\begin{tabular}{ll}
\midrule
\textbf{Paper} & \textbf{Content}            \\ \midrule
Purcell - 2023 \cite{DTAgRev1}       & Agricultural applications            \\ \addlinespace
Nie - 2022 \cite{DTAgRev2}           & AI in crop development processes        \\ \addlinespace
Attaran - 2023 \cite{DTAgRev3}       & Industry applications         \\ \addlinespace
Nasirahmadi - 2022 \cite{DTAgRev4}   & Digitization and Sensor Data  \\ \addlinespace
Pelardinos - 2023 \cite{DTAgRev5}    & Technical status-quo \\ \addlinespace
Khebbache - 2023 \cite{DTAgRev6}     & Irrigation Management \\ \addlinespace
Holzinger - 2022 \cite{DTAgRev7}     & Human-Centered AI in Agriculture
\end{tabular}
\end{table}

\noindent
Purcell \& Neubauer \cite{DTAgRev1} conduct a review on Digital Twins in agriculture and conclude that current applications benefit from the start-of-the-art technologies such as IoT, Machine Learning and Cyber-Physical Systems. However, current research is limited and focused on proving feasibility and novel methods must be adopted to apply the concept to all agricultural use cases.\\

\noindent
Nie et al.\cite{DTAgRev2} propose dividing the crop growing process into preproduction, mid-production, and postproduction stages, emphasising the use of artificial intelligence (AI) in each phase. Despite the current use of AI, the authors highlight the need for further testing of methods, analysis of existing approaches in theory and practice, and addressing issues related to data acquisition, storage, safety, and cost in future developments. In the context of Digital Twins, the authors note that most applications focus on a single entity, such as a plant or an animal, and advocate for standardisations across different applications as well as larger-scale implementations of Digital Twins.\\

\noindent
Attaran \& Celik \cite{DTAgRev3} review a broader range of applications for Digital Twins and discuss agricultural use cases as one application area. The publication attests that agricultural Digital Twins are in the early development stages. However, applications in efficiency and productivity optimisation, as well as weather modelling, soil management, supply chain management, and livestock monitoring, exist and are being further developed.\\

\noindent
Nasirahmadi \& Hensel \cite{DTAgRev4} concentrate their review on soil and irrigation management, along with Digital Twin applications for farming machinery and post-harvest processes. Despite limited research in this domain, the findings suggest promising research avenues include optimising processes, predicting optimal management decisions, and monitoring and maintaining machinery.\\

\noindent
Pelardinos et al. \cite{DTAgRev5} point out the low number of agricultural applications in Digital Twin research. Most Digital Twins predominantly rely on simulation, while many other initiatives remain conceptual, necessitating further research. The significance of sensors in (IoT)-based Digital Twins is emphasised, and cloud-based services are identified to handle the increasing volume of sensor data best. Visual model-based Digital Twins leverage game engines for 3D representations of real-world entities. The authors stress the need for enhanced focus on developing reference models and case studies in future research. They identify the current state of IoT technologies as a constraint for accurate Digital Twin models, advocating for broader applications integrating 3D visualisation, augmented reality (AR), virtual reality (VR), and geographic information systems (GIS). In agriculture, the potential extends beyond plant representations, urging a holistic focus on entire farms.\\

\noindent
Khebbache et al. \cite{DTAgRev6} argue that the amount of literature on smart irrigation has increased in recent years due to the increased use of IoT. While machine learning and deep learning are currently the primary methods for solving irrigation system tasks, Digital Twins play a minor role in representing irrigation systems but could potentially be used for sensor monitoring, presenting live readings of sensor data and predicting sensor failures. Future developments related to Digital Twins should extend beyond water management and encompass the entire soil by mapping and modelling it comprehensively.\\

\noindent
Holzinger et al. \cite{DTAgRev7} highlight the importance of explainable and robust AI in agriculture and forestry, as they are crucial to human life. AI applications are classified as autonomous, automated, assisted, or augmenting and examples in agriculture and forestry are given. In general, trust in AI decisions can be increased by incorporating human/expert knowledge. Various challenges related to technical implementations, automated systems, and robotics, as well as improving farmers' access to AI, are discussed, with the goal of Digital Twins being to represent entire farm systems virtually and include expert knowledge.

\subsection{Machine Learning in agriculture}
\noindent
Machine Learning has become a topic of interest in many research areas, and agriculture is no exception, as is exemplified by the four review papers in Table~\hyperref[tab:RW_2]{\ref{tab:RW_2}}. With Machine Learning being a wide-ranging topic, the models used in agriculture start at simple regression and classification tasks but also include various neural network structures that are used, for example, in task automation. The diverse agricultural use cases are summarised in this section.

\begin{table}[ht]
\caption{\label{tab:RW_2} Corpus of papers reviewed in Section 4.2}
\begin{tabular}{ll}
\midrule
\textbf{Paper} & \textbf{Content}            \\ \midrule
Gautron - 2022 \cite{MLAgRev1}    & RL for crop management \\ \addlinespace
Benos - 2021 \cite{MLAgRev2}      & Management of agricultural systems        \\ \addlinespace
Sharma - 2020 \cite{MLAgRev3}     & Prediction of agricultural systems   \\ \addlinespace
Abioye - 2023 \cite{MLAgRev4}     & ML in irrigation Management 
\end{tabular}
\end{table}
\noindent
According to Gautron et al. \cite{MLAgRev1}, reinforcement learning shows promise as a technique for decision support in crop management tasks as it learns from real-world experiments. However, its applications have been limited due to varying user goals for different tasks, limited data availability, and the high risks associated with taking wrong actions, particularly concerning food security. Theoretical challenges in this field include efficient learning, modelling decision problems, creating explainable policies, and handling multiple objectives under resource constraints. One possible solution for these challenges is the multi-armed bandit framework.\\

\noindent
Benos et al. \cite{MLAgRev2} conducted a comprehensive review of machine learning applications in crop, water, soil, and livestock management, focusing on crop management. Remote sensing image data is commonly utilised, and ANNs and ensemble learning are deemed the most efficient models. The integration of machine learning with Information and Communication Technology is seen as a solution to future agricultural challenges. Decision Support Systems tailored to specific cultivation systems utilise collected data, promoting sustainable and productive farming. Nevertheless, the upfront costs for farmers must be acknowledged and mitigated when implementing these systems, particularly in developing economies.\\

\noindent
Sharma et al. \cite{MLAgRev3} concluded that regression tasks are most common for predicting soil properties, weather, and crop yield, while deep learning is more frequently used in classification tasks such as pest detection. Automating
harvesting or fertilisation tasks by AI-empowered robots or drones can help complete work more efficiently. To successfully deploy smart systems for every farmer, addressing challenges related to improving model performance, educating and motivating farmers, and addressing connectivity issues in rural areas is essential.\\

\noindent
Abioye et al. \cite{MLAgRev4} used various techniques, from simpler models such as k-means to advanced methods such as RNNs, CNNs and reinforcement learning for autonomous irrigation. Challenges included limited data set availability, limited access to cloud services, and the high cost of digitising farms and infrastructure. Proposed future research emphasises reinforcement learning's adaptability and self-learning, federated learning for enhanced data security, deploying technologies in less developed countries, and exploring Digital Twins for smart irrigation. The role of fertigation in generating training data is recognised.

\section{Current Applications and Future Directions}
\begin{figure}[ht]
\begin{center}
\includegraphics[width=0.5\textwidth]{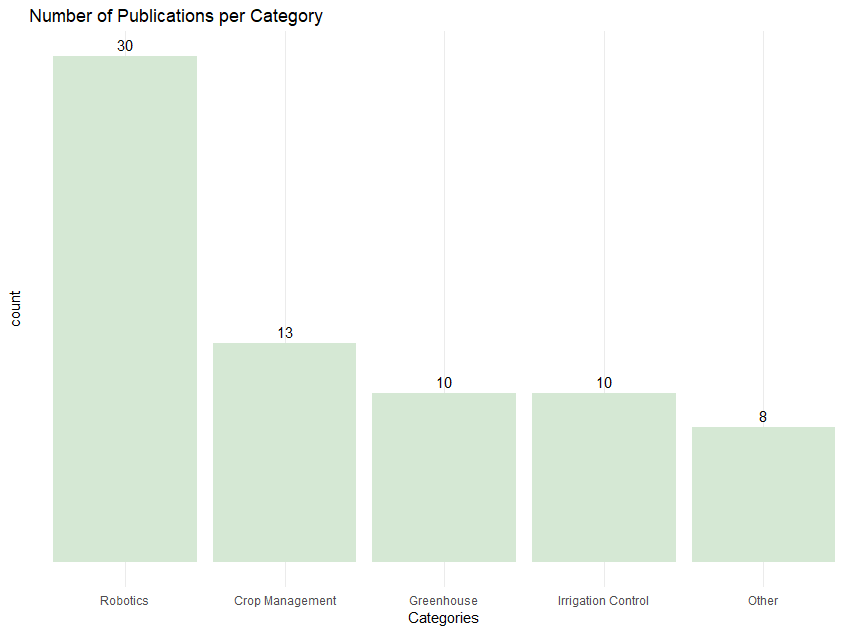}
\caption{Publications per Category}
\label{fig:Clusters_barchart}
\end{center}
\end{figure}

\begin{figure}[ht]
\begin{center}
\includegraphics[width=0.5\textwidth]{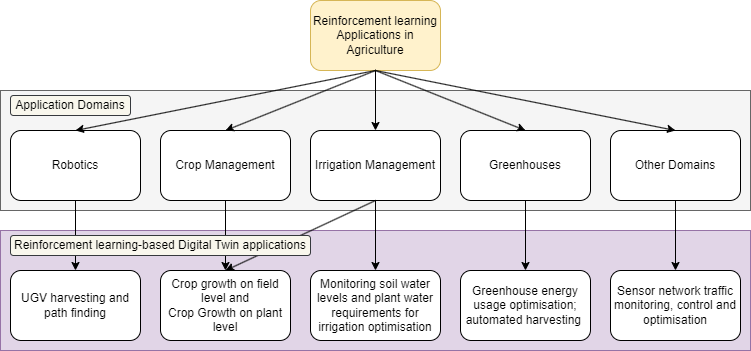}
\caption{Reinforcement learning-based Digital Twin applications for each Category}
\label{fig:Clusters_appl}
\end{center}
\end{figure}

\noindent
As is evident from the publications in the related work section, while theoretical interest in Digital Twins in agriculture is substantial, practical implementation remains limited. Current research focuses on feasibility rather than broad adoption across diverse agricultural sectors. Conversely, various machine learning methodologies to solve problems in agriculture are used, showcasing a broad range of applications. Despite this, Digital Twins are underutilised, indicating untapped potential. Thus, exploring existing reinforcement learning applications is crucial to expanding the utility of Digital Twins in agriculture. Leveraging reinforcement learning's adaptability and autonomous learning capabilities alongside other machine learning techniques offers avenues for comprehensive agricultural automation. The simulation aspect inherent in both Digital Twins and reinforcement learning further emphasises their compatibility and potential synergy in agricultural applications.\\
To get a more accessible overview of possible topics for reinforcement learning-based Digital Twins, the potential application areas are categorised according to already existing reinforcement learning applications in agriculture. Furthermore, these applications' specific reinforcement learning techniques are categorised separately to determine the most promising technical implementations.\\
In total, 71 publications of the 200 sampled were deemed appropriate according to the criteria mentioned in the methodology section. The resulting corpus is categorised according to the area of application, which is summarised in Figure~\hyperref[fig:Clusters_barchart]{\ref{fig:Clusters_barchart}}. In Figure~\hyperref[fig:Clusters_appl]{\ref{fig:Clusters_appl}} potential reinforcement learning-based Digital Twin applications, which are described in the following sections, for each domain are summarised. A structured overview of the reviewed literature can be found in Table~\hyperref[tab:Corpus_Total]{\ref{tab:Corpus_Total}} that is attached in the Appendix section. In this table, the papers are ordered according to the category they belong to. Furthermore, the paper's topic is summarised briefly, and the reinforcement learning technique, as well as whether an actor-critic method was used or not, is listed.


\subsection{Robotics}
\noindent
The first, and by far the most prevalent area of research in recent years, is robotics. In the context of this review, robotics encompasses research on unmanned vehicles and automated machines that replicate human actions like robotic arms.\\
Within this category, unmanned aerial vehicles (UAVs), usually drones, are mentioned in most publications. Automated drones play an increasingly important role in agriculture because they offer a cheap, easy and fast way to monitor larger land areas and optimise the efficiency of farming operations. In many cases, drones are used for monitoring purposes like pest, disease, fire detection \cite{robo4}, plant growth monitoring, fly trap inspection \cite{robo3}, or localisation of autonomous vehicles in unknown terrain to ensure their connectivity to a central controlling point \cite{robo6}. To accurately execute the monitoring tasks, drone usage in agriculture is often closely linked to image processing \cite{robo1}\cite{robo2}. Aside from monitoring, in one case, a reinforcement learning agent learned to control the drone's velocity and height and spray an appropriate amount of pesticide on plants \cite{robo5}. The learning goals for these tasks are to collect and send enough data to cover the entire area of interest, avoid obstacles and, in the spraying case, spray the correct parts of the plants.\\
While drones have significantly improved the efficiency of monitoring tasks in agriculture, they are not without limitations. For instance, their battery life and data storage and submission capabilities are restricted. Therefore, agents controlling drones need to find paths efficiently, avoid obstacles and coordinate with other agents to ensure the longevity and optimal usage of the UAVs, especially in the case where more than one drone gets used at the same time \cite{robo7}\cite{robo8}\cite{robo9}\cite{robo10}\cite{robo11}. Managing battery and securing data quality levels from the technical side includes task offloading, where the agent decides whether a task should be performed by the drone or in a central data collection point, and buffer overflow and channel fading reduction. These challenges can be mitigated by controlling the speed of drones and managing their connection \cite{robo12}\cite{robo13}.\\
In an attempt to standardise reinforcement learning for UAV operations, \cite{robo14} present the OmniDrones environment, in which different agents and implementations can be tested virtually.\\
Regarding reinforcement learning methods, deep Q-learning or deep Q-networks(DQN) are the most common techniques, still \cite{robo4} \cite{robo6} and \cite{robo8} use classical tabular Q-learning models. In some cases, the learning methods are extended to a multi-agent (MARL) setting or actor-critic setting like Deep Deterministic Policy Gradient (DDPG) to ensure better learning results \cite{robo6}\cite{robo7}\cite{robo9}\cite{robo13}. Even though Q-learning is at the core of most learning strategies, the approaches differ in their environment exploration strategies and overall goals, critical components for efficiently learning optimal policies. \cite{robo1} use the k-nearest neighbour (KNN) algorithm to cluster similar states and use the difference between the clustering results and a convolutional neural network (CNN) based crop health prediction to determine the flight direction. \cite{robo2} tackle a task scheduling and a crop monitoring problem. In the task scheduling case, the agent must decide whether to solve a task on edge, cloud, or FoG devices. Due to time constraints, computing an optimal solution is infeasible, and therefore, the agent relies on the ant colony optimisation heuristic (ACO) to schedule the tasks appropriately. A classical DQN performs the task of prediction and monitoring in the cloud or fog. To help their DQN implementation learn optimal strategies faster \cite{robo3} use rapidly exploring random trees (RRT) for quicker environment exploration.\\
\newline
Robots hold the potential to revolutionise the agriculture industry by taking over labour-intensive tasks such as harvesting and fruit picking, which are currently costly and time-consuming. Current research focuses on automating robotic arms to perform these tasks. The ultimate goal for these fruit-picking robots is to identify the fruit accurately, chart the optimal path to it while avoiding obstacles, and harvest it without causing any damage to the fruit or the plant \cite{robo15}\cite{robo16}\cite{robo17}. The agent controlling the arm should also be able to plan the order of picking the fruits, which can be achieved by exploring the environment and selecting the next best target \cite{robo18}. Case studies have demonstrated the successful application of these robots in apple detection and picking, cherry tomato picking, and automated harvesting in greenhouses \cite{robo19}\cite{robo20}\cite{robo21}. 
Due to their linear layout, vineyards are the testing ground for moving robots that, in the first step, learn how to reach the end of the line. In the future, these robots will also learn to monitor, spray and harvest grapes \cite{robo22}. Like harvesting, crop pruning is considered a labour-intensive task necessary to ensure crop health and maintain crop yields. One publication presented a reinforcement learning agent that learns to prune vine crops \cite{robo23}.\\
For harvesting tasks of larger fruits like bananas or fruits that are more challenging to reach, heavier machinery than a robot arm is needed \cite{robo24}\cite{robo25}. Current research for these unmanned ground vehicles (UGVs) is at a similar stage to that of UAVs. Pathfinding for harvesting machines or tractors on fields is necessary to cover the entire area and efficiently fulfil a given task. Furthermore, obstacles and avoiding them present the same problems on the ground and in the air. Approaches to pathfinding include presenting the UGV with a topographical map of the land in advance, planning a path according to it, or using sensors and cameras to monitor the area close to the vehicle and decide, given on the sensory input, where to drive \cite{robo26}\cite{robo27}\cite{robo28}\cite{robo30}. An additional challenge for ground-based vehicles is moving in difficult terrain, as the soil in fields is uneven or the fields are located in undulated terrain. Therefore, the agent must learn how much power to use and which wheels to power so as not to damage the vehicle or the soil \cite{robo29}.\\
The technical reinforcement learning implementations for robotic arms and UGVs are very similar to those used for UAVs, with DQN being used most frequently. Critic-based methods are also as prevalent and range from soft actor-critic (SAC) methods \cite{robo20}\cite{robo22}\cite{robo26}\cite{robo30} to custom implementations of student-teacher relations. In the last case, the teacher network acts as a target network with additional information not present in the student network and, at the same time, acts as a critic to the student \cite{robo17}. Interstingly, experience replay for faster and more stable learning is more prevalent for UGVs and robotic arms \cite{robo18}\cite{robo20}\cite{robo25}. Regarding reward signals, \cite{robo24} model their banana harvesting problem in a sparse reward setting, meaning that the agent rarely observers positive rewards. Therefore, the authors implement an automatic goal generation, randomly sampling targets along the way to the overall goal to facilitate efficient learning. To solve the pathfinding problem, \cite{robo28} gradually increase the reward in circular areas around the target.\\
\newline
As is evident from the abundance of literature assigned to this category, reinforcement learning delivers promising results regarding the automatisation of UAVs or UGVs. The clearly defined goals, such as monitoring a predefined area, picking a fruit or reaching a targeted area by driving there, help define action spaces and rewards, making it easy to define a reinforcement learning problem for these tasks. More challenging is the accurate representation of the environment to achieve satisfying learning results and, therefore, the definition of the state space. Three-dimensional environments often require lots of computing power in their creation, and it takes lots of effort to represent the robot and the target accurately.\\
Especially compared to crop management tasks, predefined environments to quickly implement and test an agent are scarce, leaving lots of possible development paths in that regard.\\
In robotic automation, reinforcement learning seems to be the most promising machine learning approach compared to other techniques, as it is self-learning and does not require a predefined dataset. For pathfinding problems where obstacle avoidance is the primary goal, heuristic approaches such as ant colony optimisation might be a competitor to reinforcement learning techniques that deliver similar results where, in addition, the decision-making process can be explained.\\
Concerning reinforcement learning-based Digital Twins, UGV applications are particularly promising domains because of the clearly defined real-world entity to model and goals. Therefore, monitoring the vehicle's condition and the surrounding area can be connected to learning policies for automated task completion. Furthermore, the agent's behaviour under changing conditions, for example, if different crops are planted or the terrain changes, can be simulated and allows the assessment of potential scenarios before actually employing the vehicle in the real world. 

\subsection{Crop Management}
\noindent
The second largest category contains publications that present developments in crop management practices. For this paper, crop management encompasses research in crop yield prediction, nutrition management, crop growth estimation and crop planning. Due to the comparatively large number of publications, irrigation management and greenhouse applications will form separate categories.\\
In current research on reinforcement learning for crop management, the overarching goal is to maintain, increase and predict crop yields. Short and long-term weather patterns, soil conditions and plant conditions directly influence crop yields, so the main challenge is to find how and to what extent each factor influences crop yields. As no linear or non-linear relationship could be modelled, \cite{croma1} used reinforcement learning to solve the crop yield prediction problem. Due to the predictor variables' time dependence, a recurrent neural network (RNN) is used to predict crop yields and initialise the weights of the reinforcement learning agents DQN. The agent uses the RNN prediction as a target and tries to achieve the goal by selecting parameter values for each variable to attain the predicted value. Another approach to handle the non-linear relationship is to use a reinforcement forest that was developed by \cite{croma2}. In random forests, much focus lies on the splitting criterion of a node. In a reinforcement forest, a reinforcement learning model determines the variables' importance in each node and then splits based on the computed importance value.\\
Crop growth and, therefore, crop yields are highly dependent on soil nutrient levels, and it is of great interest to measure the effects of different nutrition levels on crop development and influence them accordingly. Reinforcement learning agents can learn to manage nitrogen levels aiming at increasing crop yields \cite{croma3} or rely on the classification output of a CNN to detect malnourished rice and suggest fertilising it with nitrogen, phosphorus or potassium \cite{croma4}.\\
Other publications in crop management focus on IoT-based sensor detection for optimising crop yields. \cite{croma5} use temperature, humidity, fertiliser usage and rainfall as an input to a DQN-based agent to ensure optimal crop growth and yield, whereas \cite{croma6} focus solely on beetroot by training a DDPG-based agent to model crop development based on light intensity, temperature, CO2 levels, humidity and soil nutrient levels. \\
A novel approach to reinforcement learning is taken by \cite{croma7}, who try to predict crop evaporation based on minimal and maximal temperature and sunshine for each day. The employed reinforcement learning model learns to pick the best baseline model for each prediction timestep among CNN with long short-term memory (CNN-LSTM), Convolutional-LSTM (Conv-LSTM), CNN with eXtreme gradient boosting (CNN-XGB) and CNN with support vector regression (CNN-SVR). According to the authors, this ensemble approach is necessary because each model alone cannot predict the evaporation as well as the ensemble due to different baseline accuracies.\\
Crop planning and crop rotations can positively impact crop yields, which leads \cite{croma9} to develop a DQN-based agent that recommends crops based on soil conditions.\\
In contrast to all other categories presented in this paper, crop management sees much attention in reinforcement learning environment development. The environment is a crucial part of any reinforcement learning implementation, as the agent learns the effects of actions by interacting with the environment and observing states that the environment presents. Due to its simple usability, in general, reinforcement learning applications like simple games, the Gymnasium environment (Gym) is used to train agents and evaluate their performance \cite{GymnasiumEnv}. Efforts in agriculture have been undertaken to create environments that emulate the Gym structure but make it usable for agricultural simulations. Available Gym-based environments are CropGym \cite{croma11}\cite{croma12} that incorporates multiple process-based plant growth models and allows agents to study the effects of different nitrogen fertilisation schemes and CyclesGym \cite{croma13} where agents can learn crop rotation policies based on soil nitrogen levels and simulate plant growth among factors like soil nitrogen, carbon levels water balance and external perturbations. Another environment for a simple plant simulation model that also has a Gym-based interface was implemented by \cite{croma9}.\\
An approach outside the Gym-based environment realm is realised by \cite{croma10}, who developed an environment around the soil and water assessment tool (SWAT). Reinforcement learning agents trained within this environment learn to optimise crop yields while reducing water and fertiliser usage and, therefore, saving resources but keeping track of external factors like temperature, soil moisture levels and precipitation.\\
\newline
Many crop management tasks involve predicting future outcomes, like crop yields, crop evaporation or crop growth. Frequently, many factors that cannot be modelled in a linear or non-linear fashion influence these prediction targets. Through its exploration property, reinforcement learning learns how the prediction target reacts if external factors are changed and can find suitable predictions. Another advantage of using reinforcement learning for crop management is that much effort has already been made towards standardising environments, allowing faster development of new agents.\\
Usually, for prediction tasks, there are historical data and data on influencing factors available. Especially if the assumption of a linear relationship is reasonable, the available data enables using simpler prediction models that are even more accessible to deploy than new agents in predefined environments.\\
Crop modelling on a field or plant level are potential applications for reinforcement learning-based Digital Twins. These Digital Twins can help optimise crop yields for entire fields by determining optimal crop orders, considering external weather factors and managing water and fertiliser usage. For closer inspection of how these factors influence a particular plant and how it would react to changes in these conditions, the smaller-scale plant-level Digital Twins could be used.  

\subsection{Irrigation Management}
\noindent
An essential point of every agricultural operation is irrigation management, as plant growth and crop yields are heavily influenced by the amount of water they get. Furthermore, water usage in agriculture is a topic that sees lots of interest in the context of climate change and more frequent occurrences of water shortages worldwide. Because it is such an important topic, there are many attempts at reinforcement learning controlled irrigation management in agriculture. Compared to the topics in the other sections, the goals for publications within this category are similar: keep crops healthy by controlling soil moisture levels while using as little water as possible. Aside from the crop's regular water usage that it needs for growth, weather patterns and evaporation are factors that influence soil moisture levels. \cite{water1}\cite{water3}. Specific crops that were used for testing irrigation schemes are tomatoes, rice and maise \cite{water4}\cite{water5}\cite{water6}\cite{water7}. Greenhouses lend themselves well to research as they present a controlled environment, which will be discussed more closely in the next section. However, \cite{water8} specifically focus on greenhouses' irrigation management. \cite{water9} train multiple DQN-based agents for various plots of land and observe the effects of water usage in the Colorado River Basin according to the Colorado River Simulation System.\\
Similarly to crop management in general, environments to train reinforcement learning agents are also developed for irrigation management. Both the Aquacrop-gym and the gym-DSSAT environments are Gym-based environments that can be used to develop irrigation policies \cite{water2}\cite{water10}. The gym-DSSAT implementation transfers the frequently used DSSAT model into a Python environment.\\
Every reinforcement learning agent developed for use-cases in this category relies on DQN, except for the one used by \cite{water1}, who use a tabular Q-learning approach and \cite{water3}, who implement Proximal Policy Optimization (PPO). \cite{water4}\cite{water5} and \cite{water7} employ actor-critic methods with \cite{water4} using an LSTM network and CNN for yield predictions in their tomato case study.\\
\newline
Especially in the context of environmental sustainability, water management requires lots of precision in order not to waste unnecessary resources. As the required water for optimal crop growth can be monitored through soil moisture, crop evaporation and natural precipitation, reinforcement learning agents can quickly assess the situation and decide when to water the plants dynamically. Using the available sensor data also allows to virtually replicate the entire water management system, which can be integrated into the field-level reinforcement learning-based Digital Twins discussed in the previous section.\\
The required data to assess the current moisture levels can also be used for other heuristic or non-heuristic decision support algorithms that suggest when and how much water to use. Given the importance of saving water amidst climate change, wasting as little water as possible should be the primary goal in irrigation management. Quick adaptions to changing conditions are required and should be an optimal use case for reinforcement learning agents.

\subsection{Greenhouses}
\noindent
Compared to regular farms, greenhouses allow control of environmental parameters like temperature, light, precipitation, and humidity. The ability to control for external factors in plant growth makes greenhouses an interesting testing ground for researchers and makes it possible to grow plants all year round. However, due to the constant environmental control, greenhouses use lots of energy, making them less sustainable compared to farming outside greenhouses.\\
Since automated controlling is one of the strengths of reinforcement learning, it is no surprise that agents are being developed to control the environment efficiently in greenhouses with the push towards more sustainable farming practices. Different from the applications in the other categories, there are more common models for greenhouse reinforcement learning agents than DQN alone. In most cases, actor-critic approaches like DDPG \cite{green1} and SAC \cite{green2} are used. For \cite{green2} and \cite{green6}, avoiding worst-case scenarios is essential, and they achieve this by masking specific actions that lead to critical conditions in advance. \cite{green3} use a MARL approach to simulate climate control by splitting the action space into subactions, building a reinforcement learning model for each subaction and maintaining correlations between the different actions. These submodels are embedded into a hierarchical structure, where one overarching model controls the submodels. The framework is called: structured cooperative reinforcement learning algorithm (SCORE). Another multi-agent reinforcement learning model is implemented by \cite{green4}, who aim to stabilise the power usage of their greenhouse to reduce stress on the power network while also trying to reduce the total power consumption. The agents are connected by a shared attention mechanism to facilitate faster learning and better cooperation between the agents. To avoid online learning methods, \cite{green5} rely on historical data and climate trajectories of their greenhouse DQN. In addition to the agent, a mixed integer linear program is defined after each learning phase that helps the agent avoid overfitting to the historical data.\\
In contrast, \cite{green7} use their greenhouse's available climate and plant-specific sensor data to control the artificial lighting using a DQN agent dynamically. Controlling artificial lighting influences not only energy consumption but also plant growth, which \cite{green8} use to increase the dry mass of Spirulina Sp. An LSTM network was used to predict the next day's light intensity by relying on data from the past days, and based on the prediction, a tabular Q-learning model was trained to choose among four options how much light should be artificially added to maximise the dry mass of Spirulina Sp.\\
Most greenhouse applications presented so far rely on sensor data at some point in their automatisation process. That means the quality of their models depends on the quality of the collected data, which is why it is crucial to determine optimal sensor locations within a greenhouse. \cite{green9} try to find an optimal solution to the sensor placement problem by implementing a Bayesian reinforcement learning approach that relies on Thompson sampling for exploration and exploitation. In addition to climate control, crop yield predictions, for example for strawberries, often grown in greenhouses due to their permanent demand, can be made using the available climate data. The available data is fed into an informer that predicts strawberry growth. The predicted values are then presented as a target to a Q-learning model that learns to regulate climate conditions by aiming to achieve the target \cite{green10}.\\
\newline
As energy efficiency for greenhouses is one of the most pressing goals for research in this category, reinforcement learning shows its strength in simultaneously handling multiple variables such as temperature, humidity, and light intensity to control the greenhouse climate accurately. With respect to reinforcement learning-based Digital Twins, greenhouses behave similarly to UGVs in 
that there is a clearly defined entity to replicate and the optimisation goals for reinforcement learning can also be defined in straightforward way. Again the simulation aspect of Digital Twins allows to gain knowledge how certain actions and conditions would affect the greenhouse in advance and therefore help identify potential improvements for their energy consumption.\\
Besides energy reduction, the goal is maintaining an environment that secures stable crop yields or even increases them. However, depending on the available data, the problem might also be solved using a linear program if constraints to variables and their relation to crop development are known. Other techniques that can also handle non-numeric variables and are able to suggest actions are decision tree-based methods. Both linear programs and decision tree methods are more straightforward to implement, and the performed actions are easier to understand. However, reinforcement learning agents might react faster to changing conditions and, therefore, handle the climate more accurately.

\subsection{Other Applications}
\noindent
The publications in this category do not fit into any of the previous groups. However, this does not imply that the works' contribution can be considered entirely independent of the previously discussed topics. Many applications rely on collected sensor data, and \cite{o1}, \cite{o2} and \cite{o3} illustrate how reinforcement learning manages data transmission. The publications focus on the technical solution to the transmission problem, which has potential applications in agriculture and beyond. According to \cite{o1}, sensors are small, light, and inexpensive but have a relatively short battery life. Therefore, the DDPG algorithm is used to learn a policy that aims to send information regularly to keep it up to date while trying to conserve energy simultaneously. The agent collects rewards only if the information is accurate and the time between two transmissions is long.\\
\cite{o2} also aim to extend the battery life of sensors like UAVs, but they focus less on the accuracy of data transfer than on solving the tasks efficiently. To achieve this, the authors have implemented a Q-learning agent trained to maximise the drone's battery life while simultaneously finishing given tasks before a deadline. The reinforcement learning model for each drone can determine whether it completes the task, passes it on to another drone, or sends it to a shared data collection point for it to be finished there.\\
Efficient data transmission is crucial for sustainable smart farming. Smart farms typically use multiple sensors distributed over a large area and consume significant amounts of energy. To optimise data transmission between sensors and a central data collection point, \cite{o3} developed a tabular Q-learning model to identify the optimal transmission path. The agent aims to achieve complete and energy-efficient transmissions and transmit data without delay, even in areas without signal. The authors claim that the approach can also be applied to more extensive sensor networks.\\
\cite{o4} focus on the quality of experience of video data, which can be used for general surveillance purposes and specifically for livestock monitoring in agriculture. An average advantage actor-critic (A3C) reinforcement learning agent is trained to handle dynamic network conditions by adapting the bitrates to ensure a high-quality video stream.\\
While efficient data collection and transmission are essential in optimising farming operations, it is necessary to have the means to store and process the available data securely. The most common practice in data analysis is to collect all available information in one dataset or database and access it when needed, which may require lots of data transmissions and, therefore, potential security risks. One way to mitigate this risk is to develop a federated learning network, where data analysis is done locally at the data collection point, and only model parameters are shared between different analysis entities. Deciding which user's model parameters to incorporate into the overall model is necessary to ensure model quality and accuracy. \cite{o5} propose to perform a spectral clustering on all users in a federated learning network and have a DQN-based agent decide which clusters to use for the overarching model.\\
Reinforcement learning in agriculture is not only used to increase crop yields, reduce labour or make tasks more efficient, instead \cite{o6} show that both tabular Q-learning and DQN can be used to optimise the agri-food supply chain by modelling it as a blockchain environment. The main goal for their agents is to increase product profits for the farmers.\\
Another DQN application in agriculture is frost forecasting. A Fuzzy-based DQN uses historical data to predict frosts to ensure agricultural productivity and maintain stable crop prices \cite{o7}.\\
In a very different application to frost prediction, DQNs' flexible use cases are highlighted as they can also be used for genomic selection for plant breeding parent selection. The agents' goal is to allocate resources for each plant generation and, in doing so, decide which parents are best suitable for the desired breeding goal \cite{o8}.\\
\newline
The reinforcement learning applications in this category show that using machine learning, data transmission accuracy, quality and security, as well as task offloading, can be improved. Reinforcement learning-based Digital Twins' primary use case might be to monitor and control network traffic and optimise it by more efficient task scheduling or sensor placement.\\
As discussed for the previous categories, the reinforcement learning agents might be replaced by similarly performing pathfinding heuristics or task scheduling techniques. However, technical improvements in longer-lasting batteries, more stable networks, and increased sensor transmission capabilities might outpace the quality of machine-learning implementations. Therefore, these novelties will alleviate some pressure for optimisation, as fewer sensors that last longer and transmit with a higher degree of certainty might be introduced.

\subsection{Reinforcement Learning Techniques}
\begin{figure}[ht]
\begin{center}
\includegraphics[width=0.5\textwidth]{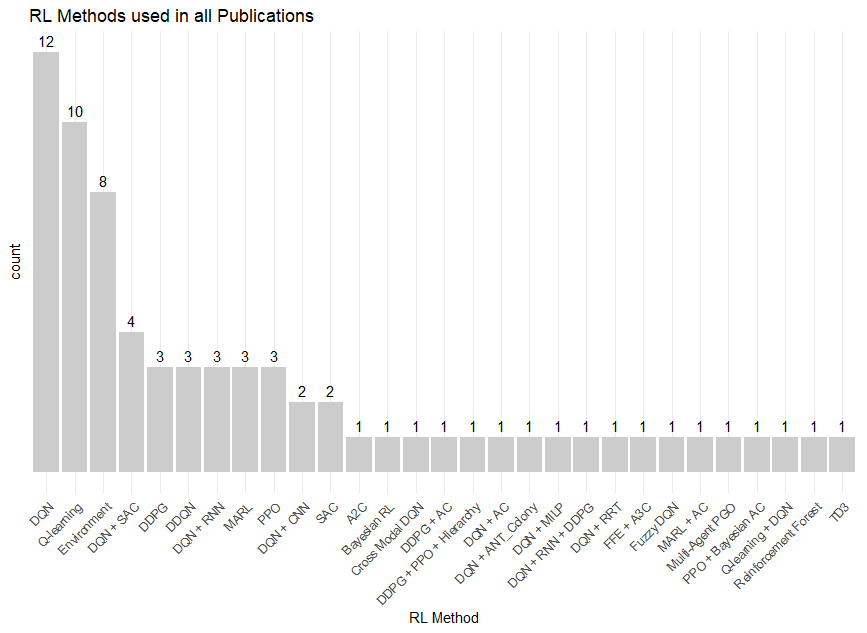}
\caption{Reinforcement Learning Techniques}
\label{fig:RL_Tech}
\end{center}
\end{figure}

\begin{figure}[ht]
\begin{center}
\includegraphics[width=0.5\textwidth]{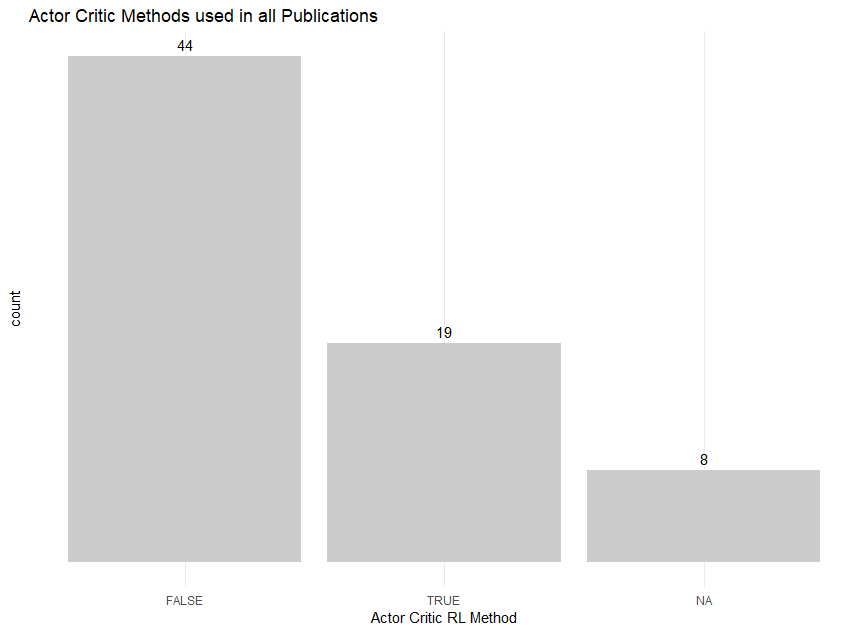}
\caption{Reinforcement Learning Techniques}
\label{fig:AC}
\end{center}
\end{figure}

\noindent
In the previous sections, different reinforcement learning methods have been presented for varying use cases. Still, a compact summary of the different models allows to assess the most prevalent techniques in current agricultural research. The reinforcement learning methods are summarised in Figure~\hyperref[fig:RL_Tech]{\ref{fig:RL_Tech}}.
DQN is by far the most common approach to solving reinforcement learning tasks. As described in the previous sections, DQN is suitable for many tasks, such as energy control in greenhouses, controlling a robotic arm or recommending crop planting orders and irrigation schemes. Frequently, though, DQNs are combined with other techniques to facilitate faster learning by efficiently exploring the state space or collecting higher rewards, yielding better policies. In time series-dependent situations, RNNs are commonly connected to DQN agents as they allow the retention of a memory of past events. For image processing tasks, CNNs have proven reliable in, for example, outlier detection, where the agent decides how to best act given an outlier. Actor-critic methods such as SAC are used to optimise the learned policies. Figure~\hyperref[fig:AC]{\ref{fig:AC}} summarises in how many publications actor-critic methods were used in total. Missing values signify that a learning environment was created that allows the implementation of various reinforcement learning methods, both actor-critic and non-actor-critic.
In addition, for specific tasks, DQNs were combined with a Fuzzy learning approach, ant colony optimisation, and a multi-integer linear program, or they were embedded in a hierarchical structure like a tree or used multi-agent settings.\\
Multi-agent (MARL) settings are standard for cases where multiple entities like a swarm of drones or different water users need to be controlled simultaneously or if an agent handles the subdivisions of state or action spaces.\\
Even though deep reinforcement learning methods are the most common, there are cases where tabular methods like tabular Q-learning are used, especially if the action space is small enough. Tabular methods usually require more memory space as tables for state-action pairs must be maintained. However, their easy implementation and traceable decision-making processes still make them an attractive alternative to more involved deep learning methods.\\
Even though there are methods that allow for interpretations of learned decisions, the explainability of reinforcement learning algorithms for agricultural use cases is a topic that is widely disregarded. However, explainable decisions are a crucial part of gaining trust in learned policies and, therefore, help decrease the reluctance of farmers to use machine learning-powered decision support.

\section{Conclusion}
\noindent
In conclusion, this review highlights the early developmental stage of Digital Twins in agriculture but also the widespread utilisation of various machine learning techniques within the agricultural sector. Among these machine learning techniques, reinforcement learning is a frequently used approach, particularly in the defined categories of robotics, crop management, irrigation management, and greenhouses. The most common method to solve reinforcement learning problems is to use Deep Q-Networks (DQN). It was also shown that reinforcement learning is often combined with Recurrent Neural Networks (RNNs), Convolutional Neural Networks (CNNs), and other optimisation techniques. However, there is a notable absence of focus on explainable reinforcement learning techniques, indicating a significant area for future development, especially considering the importance of trustworthiness in AI.\\
Moreover, this review highlights already existing applications for reinforcement learning-based Digital Twins, particularly in robotics, greenhouses, and crop models. The scope of implementation for these domains is manageable due to the clearly defined and replicable nature of the corresponding real-world entities. Even though some of the Digital Twins are more advanced, the lack of standardised learning environments and applications only in small-scale use cases indicates that there is a need to optimise these applications further to implement more efficient Digital Twins faster. Exploring the integration of reinforcement learning-based Digital Twins in these areas presents promising opportunities to advance agricultural automation and enhance productivity.

\section*{Acknowledgements}
\noindent
The authors acknowledge the funding of the project “digital.twin.farm” funded by the Austrian Federal Ministry of Education, Science and Research.

\newpage
\onecolumn
\newpage
\appendix
\section*{Appendix}

\begin{xltabular}{\textwidth}{@{}XXXXX@{}}
    \caption{\label{tab:Corpus_Total} Corpus of the reviewed Publications}  \\
\toprule
Authors  & Topic & Category  & RL Method & Actor-Critic \\* 
\midrule
\endhead
Zhang - 2020 \cite{robo1}                 & UAV Crop Monitoring                          & Robotics           & DQN                    & FALSE        \\ \addlinespace
Ganesh - 2023 \cite{robo2}                & UAV Crop Monitoring                          & Robotics           & DQN + ANT\_Colony      & FALSE        \\ \addlinespace
Boubin - 2022 \cite{robo7}                & UAV Swarm Management                         & Robotics           & MARL                   & FALSE        \\ \addlinespace
Castro - 2023 \cite{robo3}                & UAV Path Planning and Monitoring             & Robotics           & DQN + RRT              & FALSE        \\ \addlinespace
Pamuklu - 2023 \cite{robo4}               & UAV Crop Monitoring                          & Robotics           & Q-learning             & FALSE        \\ \addlinespace
Yang - 2022 \cite{robo26}                 & UGV Path Planning                            & Robotics           & SAC                    & TRUE         \\ \addlinespace
Martini - 2022 \cite{robo22}              & UGV for Vineyard Monitoring                  & Robotics           & DQN + SAC              & TRUE         \\ \addlinespace
Hao - 2022 \cite{robo5}                   & UAV Spraying Automation                      & Robotics           & DQN + AC               & TRUE         \\ \addlinespace
Pourroostaei Ardakani - 2021 \cite{robo8} & UAV Route Planning and Monitoring            & Robotics           & Q-learning             & FALSE        \\ \addlinespace
Testi - 2020 \cite{robo6}                 & UAVs for UGV Monitoring and Connection       & Robotics           & MARL                   & FALSE        \\ \addlinespace
Guichao - 2021 \cite{robo9}               & UAV Route Planning                           & Robotics           & DQN + RNN + DDPG       & TRUE         \\ \addlinespace
Petrenko - 2020 \cite{robo21}             & Automated Harvesting Robot in Greenhouse     & Robotics           & DQN + RNN              & FALSE        \\ \addlinespace
Faryadi - 2021 \cite{robo27}              & UGVs in map field Topography                 & Robotics           & Q-learning             & FALSE        \\ \addlinespace
Yinchu - 2023 \cite{robo25}               & UGVs for Kiwifruit Picking                   & Robotics           & DQN                    & FALSE        \\ \addlinespace
Zeng - 2022 \cite{robo18}                 & UGV Automated Environment Detection          & Robotics           & DDQN                   & FALSE        \\ \addlinespace
Yajun - 2024 \cite{robo20}                & Cherry Tomato Picking                        & Robotics           & DQN + SAC              & TRUE         \\ \addlinespace
Tian - 2023 \cite{robo15}                 & Arm Path Planning for Fruit Picking          & Robotics           & Environment            &              \\ \addlinespace
Yandun - 2021 \cite{robo23}               & Vineyard Pruning                             & Robotics           & DQN + CNN              & FALSE        \\ \addlinespace
Andriyanov - 2023 \cite{robo19}           & Apple Detection and Picking                  & Robotics           & Q-learning             & FALSE        \\ \addlinespace
Wiberg - 2022 \cite{robo29}               & Heavy UGV Path Finding in Difficult Terrain  & Robotics           & PPO + Bayesian AC      & TRUE         \\ \addlinespace
Josef - 2020 \cite{robo28}                & UGV for Close Range Sensing and Path Finding & Robotics           & DQN                    & FALSE        \\ \addlinespace
Nguyen - 2022 \cite{robo12}               & UAV Task Offloading in Area Surveillance     & Robotics           & DQN                    & FALSE        \\ \addlinespace
Nethala - 2023 \cite{robo16}              & Arm Path Planning for Fruit Touching         & Robotics           & DDPG + PPO + Hierarchy & TRUE         \\ \addlinespace
Yang - 2020 \cite{robo17}                 & Vision-based Arm Learning                    & Robotics           & Cross Modal DQN        & FALSE        \\ \addlinespace
Lin - 2022 \cite{robo24}                  & Large UGV for Banana Harvesting              & Robotics           & TD3                    & TRUE         \\ \addlinespace
Kurunatahn - 2021 \cite{robo13} & UAV Cruise Control to reduce Buffer Overflows and Channel Fading & Robotics & DDPG &
  TRUE \\ \addlinespace
Alon - 2020 \cite{robo10}                 & UAV Swarm Coordination and Cooperation       & Robotics           & Multi-Agent PGO        & TRUE         \\ \addlinespace
Xu - 2024 \cite{robo14} & OmniDrones - Environment for RL Drone Implementations & Robotics & Environment &
   \\ \addlinespace
Roghair - 2022 \cite{robo11}              & UAV Obstacle detection                       & Robotics           & DQN                    & FALSE        \\ \addlinespace
Din - 2022 \cite{robo30}                  & Land Area Coverage Control                   & Robotics           & DDQN                   & FALSE        \\ \addlinespace
Elavarasan - 2020 \cite{croma1}           & Crop Yield Prediction                        & Crop Management    & DQN + RNN              & FALSE        \\ \addlinespace
Wu - 2022 \cite{croma3}                   & Nitrogen Management                          & Crop Management    & DQN + SAC              & TRUE         \\ \addlinespace
Yassine - 2022 \cite{croma5}              & Crop Growth and Yield Management             & Crop Management    & DQN + RNN              & FALSE        \\ \addlinespace
Madondo - 2023 \cite{croma10}             & Yield Maximization and Resource Reduction    & Crop Management    & Environment            &              \\ \addlinespace
Zheng - 2021 \cite{croma6}                & Sensor-based Crop Yield Increase             & Crop Management    & DDPG + AC              & TRUE         \\ \addlinespace
Sharma - 2022 \cite{croma7}               & Crop Evaporation Estimation                  & Crop Management    & DQN                    & FALSE        \\ \addlinespace
Elavarasan - 2021 \cite{croma2}           & Crop Yield Prediction                        & Crop Management    & Reinforcement Forest   & FALSE        \\ \addlinespace
Overweg - 2021 \cite{croma11}             & CropGym - Crop Growth Simulation             & Crop Management    & Environment            &              \\ \addlinespace
Wang - 2021 \cite{croma4}                 & Malnutrition Detection in Rice               & Crop Management    & Q-learning             & FALSE        \\ \addlinespace
Kallenberg - 2023 \cite{croma12}          & CropGym - Crop Growth Simulation             & Crop Management    & Environment            &              \\ \addlinespace
Turchetta - 2022 \cite{croma13} &
  CyclesGym - Crop Rotation Planning and Plant Growth &
  Crop Management &
  Environment &
   \\ \addlinespace
Ashcraft - 2021 \cite{croma9} &
  Crop Yield Optimisation based on Plant Growth model &
  Crop Management &
  PPO & 
  FALSE \\ \addlinespace
Bouni - 2022 \cite{croma8}                & Crop Planting Recommendations                & Crop Management    & DQN                    & FALSE        \\ \addlinespace
Tropea - 2022 \cite{water1}               & Irrigation Management                        & Irrigation Control & Q-learning             & FALSE        \\ \addlinespace
Zhou - 2020 \cite{water8}                 & Irrigation Management in Greenhouse          & Irrigation Control & DQN                    & FALSE        \\ \addlinespace
Alibabaei - 2022 \cite{water4}            & Irrigation Management on Tomato Field        & Irrigation Control & DQN + CNN              & FALSE        \\ \addlinespace
Alibabaei - 2022 \cite{water5}            & Irrigation Management                        & Irrigation Control & A2C                    & TRUE         \\ \addlinespace
Chen - 2021 \cite{water6}                 & Irrigation Management for Rice               & Irrigation Control & DQN                    & FALSE        \\ \addlinespace
Kelly - 2024 \cite{water2}                & Irrigation Management                        & Irrigation Control & Environment            &              \\ \addlinespace
Gautron - 2022 \cite{water10}             & gym-DSSAT for Python                         & Irrigation Control & Environment            &              \\ \addlinespace
Hung - 2021 \cite{water9}                 & Water User Coordination                      & Irrigation Control & DQN                    & FALSE        \\ \addlinespace
Ding - 2022 \cite{water3}                 & Irrigation Management                        & Irrigation Control & PPO                    & FALSE        \\ \addlinespace
Tao - 2022 \cite{water7}                  & Irrigation and Fertilization Management      & Irrigation Control & DQN + SAC              & TRUE         \\ \addlinespace
Wang - 2020 \cite{green1}                 & Climate Control                              & Greenhouse         & DDPG                   & TRUE         \\ \addlinespace
Zhang - 2021 \cite{green2}                & Climate Control                              & Greenhouse         & SAC                    & TRUE         \\ \addlinespace
Uyeh - 2021 \cite{green9}                 & Sensor Placement                             & Greenhouse         & Bayesian RL            & FALSE        \\ \addlinespace
Li - 2022 \cite{green3}                   & Simulated Control                            & Greenhouse         & MARL                   & FALSE        \\ \addlinespace
Lu - 2023 \cite{green10}                  & Strawberry Yield Prediction                  & Greenhouse         & Q-learning             & FALSE        \\ \addlinespace
Ajagekar - 2024 \cite{green4}             & Power Usage Optimization                     & Greenhouse         & MARL + AC              & TRUE         \\ \addlinespace
Ajagekar - 2023 \cite{green5}             & Energy Efficient Climate Control             & Greenhouse         & DQN + MILP             & FALSE        \\ \addlinespace
Chen - 2023 \cite{green6}                 & Energy Saving through Light Control          & Greenhouse         & PPO                    & FALSE        \\ \addlinespace
Decardi-Nelson - 2023 \cite{green7} & Resource Optimization in Plant Factories through Light Control & Greenhouse & DQN &
  FALSE \\ \addlinespace
Doan - 2021 \cite{green8}                 & Light Controlled Spirulina Sp. Farming       & Greenhouse         & Q-learning             & FALSE        \\ \addlinespace
Herabad - 2022 \cite{o7}                  & Frost Forecasting                            & Other              & Fuzzy DQN              & FALSE        \\ \addlinespace
Ahmadi - 2022 \cite{o5}                   & Federated Learning user Selection            & Other              & DDQN                   & FALSE        \\ \addlinespace
Naresh - 2022 \cite{o4}                   & Video Stream Quality Improvement             & Other              & FFE + A3C              & TRUE         \\ \addlinespace
Hribar - 2022 \cite{o1}                   & Sensor Battery Life Improvement              & Other              & DDPG                   & TRUE         \\ \addlinespace
Nguyen - 2022 \cite{o2}                   & UAV Task Offloading                          & Other              & Q-learning             & FALSE        \\ \addlinespace
Moeinizade - 2022 \cite{o8}               & Genomic Selection in Plant Breeding          & Other              & DQN                    & FALSE        \\ \addlinespace
Ali - 2023 \cite{o3}                      & Sensor Battery Life Improvement              & Other              & Q-learning             & FALSE        \\ \addlinespace
Chen - 2021 \cite{o6}                     & Agri-Food Supply Chain Optimization          & Other              & Q-learning + DQN       & FALSE
\end{xltabular}

\clearpage
\twocolumn


\begin{thebibliography}{150}
\bibitem{Intro1}Malhi, G., Kaur, M. \& Kaushik, P. Impact of Climate Change on Agriculture and Its Mitigation Strategies: A Review. {\em Sustainability}. \textbf{13} (2021), \url{https://www.mdpi.com/2071-1050/13/3/1318}, Visited on 2024-01-10
\bibitem{Intro2}World Bank Group. Population Estimates And Projections.  (2024), \url{https://databank.worldbank.org/source/population-estimates-and-projections}, Accessed: 2024-01-10
\bibitem{Intro3}Burgos, D. \& Ivanov, D. Food retail supply chain resilience and the COVID-19 pandemic: A digital twin-based impact analysis and improvement directions. {\em Transportation Research Part E: Logistics And Transportation Review}. \textbf{152} pp. 102412 (2021), \url{https://www.sciencedirect.com/science/article/pii/S1366554521001794}, Accessed: 2024-01-10
\bibitem{Intro4}Lobell, D. \& Gourdji, S. The Influence of Climate Change on Global Crop Productivity. {\em Plant Physiology}. \textbf{160}, 1686-1697 (2012,10), \url{https://doi.org/10.1104/pp.112.208298}, Accessed: 2024-01-10
\bibitem{DTDef1}Brucherseifer, E., Winter, H., Mentges, A., Mühlhäuser, M. \& Hellmann, M. Digital Twin conceptual framework for improving critical infrastructure resilience. {\em At - Automatisierungstechnik}. \textbf{69}, 1062-1080 (2021), \url{https://doi.org/10.1515/auto-2021-0104}, Accessed: 2024-01-10
\bibitem{DTDef}Grieves, M. \& Vickers, J. Digital Twin: Mitigating Unpredictable, Undesirable Emergent Behavior in Complex Systems. {\em Transdisciplinary Perspectives On Complex Systems: New Findings And Approaches}. pp. 85-113 (2017), \url{https://doi.org/10.1007/978-3-319-38756-7_4}, Accessed: 2024-01-10
\bibitem{RLDef}Sutton, R. \& Barto, A. Reinforcement learning: An introduction. (MIT press,2018), Accessed: 2024-01-10
\bibitem{RLDef1}Richard Bellman Dynamic Programming. {\em Science}. \textbf{153}, 34-37 (1966), \url{https://www.science.org/doi/abs/10.1126/science.153.3731.34}
\bibitem{DTAgRev1}Purcell, W. \& Neubauer, T. Digital Twins in Agriculture: A State-of-the-art review. {\em Smart Agricultural Technology}. \textbf{3} pp. 100094 (2023), \url{https://www.sciencedirect.com/science/article/pii/S2772375522000594}, Accessed on 2024-01-11
\bibitem{DTAgRev2}Nie, J., Wang, Y., Li, Y. \& Chao, X. Artificial intelligence and digital twins in sustainable agriculture and forestry: a survey. {\em Turkish Journal Of Agriculture And Forestry}. \textbf{46}, 642-661 (2022), Accessed on 2024-01-11
\bibitem{DTAgRev3}Attaran, M. \& Celik, B. Digital Twin: Benefits, use cases, challenges, and opportunities. {\em Decision Analytics Journal}. \textbf{6} pp. 100165 (2023), \url{https://www.sciencedirect.com/science/article/pii/S277266222300005X}, Accessed on 2024-01-11
\bibitem{DTAgRev4}Nasirahmadi, A. \& Hensel, O. Toward the Next Generation of Digitalization in Agriculture Based on Digital Twin Paradigm. {\em Sensors}. \textbf{22} (2022), \url{https://www.mdpi.com/1424-8220/22/2/498}, Accessed on 2024-01-11
\bibitem{DTAgRev5}Peladarinos, N., Piromalis, D., Cheimaras, V., Tserepas, E., Munteanu, R. \& Papageorgas, P. Enhancing Smart Agriculture by Implementing Digital Twins: A Comprehensive Review. {\em Sensors}. \textbf{23} (2023), \url{https://www.mdpi.com/1424-8220/23/16/7128}, Accessed on 2024-01-11
\bibitem{DTAgRev6}Khebbache, R., Merizig, A., Rezeg, K. \& Lloret, J. The recent technological trends of smart irrigation systems in smart farming: a review. {\em International Journal Of Computing And Digital Systems}. \textbf{14}, 10317-10335 (2023)
\bibitem{DTAgRev7}Holzinger, A., Saranti, A., Angerschmid, A., Retzlaff, C., Gronauer, A., Pejakovic, V., Medel-Jimenez, F., Krexner, T., Gollob, C. \& Stampfer, K. Digital Transformation in Smart Farm and Forest Operations Needs Human-Centered AI: Challenges and Future Directions. {\em Sensors}. \textbf{22} (2022), \url{https://www.mdpi.com/1424-8220/22/8/3043}, Accessed on 2024-01-11
\bibitem{MLAgRev1}Gautron, R., Maillard, O., Preux, P., Corbeels, M. \& Sabbadin, R. Reinforcement learning for crop management support: Review, prospects and challenges. {\em Computers And Electronics In Agriculture}. \textbf{200} pp. 107182 (2022), \url{https://www.sciencedirect.com/science/article/pii/S0168169922004999}, Accessed on 2024-01-11
\bibitem{MLAgRev2}Benos, L., Tagarakis, A., Dolias, G., Berruto, R., Kateris, D. \& Bochtis, D. Machine Learning in Agriculture: A Comprehensive Updated Review. {\em Sensors}. \textbf{21} (2021), \url{https://www.mdpi.com/1424-8220/21/11/3758}, Accessed on 2024-01-11
\bibitem{MLAgRev3}Sharma, A., Jain, A., Gupta, P. \& Chowdary, V. Machine Learning Applications for Precision Agriculture: A Comprehensive Review. {\em IEEE Access}. \textbf{9} pp. 4843-4873 (2021)
\bibitem{MLAgRev4}Abioye, E., Hensel, O., Esau, T., Elijah, O., Abidin, M., Ayobami, A., Yerima, O. \& Nasirahmadi, A. Precision Irrigation Management Using Machine Learning and Digital Farming Solutions. {\em AgriEngineering}. \textbf{4}, 70-103 (2022), \url{https://www.mdpi.com/2624-7402/4/1/6}, Accessed on 2024-01-11
\bibitem{GymnasiumEnv}Foundation, F. Gymnasium An API standard for reinforcement learning with a diverse collection of reference environments. (\url{https://gymnasium.farama.org/},2023), retrieved on 09.01.2024
\bibitem{robo1}Zhang, Z., Boubin, J., Stewart, C. \& Khanal, S. Whole-Field Reinforcement Learning: A Fully Autonomous Aerial Scouting Method for Precision Agriculture. {\em Sensors}. \textbf{20} (2020), \url{https://www.mdpi.com/1424-8220/20/22/6585}
\bibitem{robo2}Devarajan, G., Nagarajan, S., T.V., R., T., V., Ghosh, U. \& Alnumay, W. DDNSAS: Deep reinforcement learning based deep Q-learning network for smart agriculture system. {\em Sustainable Computing: Informatics And Systems}. \textbf{39} pp. 100890 (2023), \url{https://www.sciencedirect.com/science/article/pii/S2210537923000458}
\bibitem{robo3}Castro, G., Berger, G., Cantieri, A., Teixeira, M., Lima, J., Pereira, A. \& Pinto, M. Adaptive Path Planning for Fusing Rapidly Exploring Random Trees and Deep Reinforcement Learning in an Agriculture Dynamic Environment UAVs. {\em Agriculture}. \textbf{13} (2023), \url{https://www.mdpi.com/2077-0472/13/2/354}
\bibitem{robo4}Pamuklu, T., Nguyen, A., Syed, A., Kennedy, W. \& Erol-Kantarci, M. IoT-Aerial Base Station Task Offloading With Risk-Sensitive Reinforcement Learning for Smart Agriculture. {\em IEEE Transactions On Green Communications And Networking}. \textbf{7}, 171-182 (2023)
\bibitem{robo5}Hao, Z., Li, X., Meng, C., Yang, W. \& Li, M. Adaptive spraying decision system for plant protection unmanned aerial vehicle based on reinforcement learning. {\em International Journal Of Agricultural And Biological Engineering}. \textbf{15}, 16-26 (2022)
\bibitem{robo6}Testi, E., Favarelli, E. \& Giorgetti, A. Reinforcement Learning for Connected Autonomous Vehicle Localization via UAVs. {\em 2020 IEEE International Workshop On Metrology For Agriculture And Forestry (MetroAgriFor)}. pp. 13-17 (2020)
\bibitem{robo7}Boubin, J., Burley, C., Han, P., Li, B., Porter, B. \& Stewart, C. MARbLE: Multi-Agent Reinforcement Learning at the Edge for Digital Agriculture. {\em 2022 IEEE/ACM 7th Symposium On Edge Computing (SEC)}. pp. 68-81 (2022)
\bibitem{robo8}Pourroostaei Ardakani, S. \& Cheshmehzangi, A. Reinforcement Learning-Enabled UAV Itinerary Planning for Remote Sensing Applications in Smart Farming. {\em Telecom}. \textbf{2}, 255-270 (2021), \url{https://www.mdpi.com/2673-4001/2/3/17}
\bibitem{robo9}Lin, G., Zhu, L., Li, J., Zou, X. \& Tang, Y. Collision-free path planning for a guava-harvesting robot based on recurrent deep reinforcement learning. {\em Computers And Electronics In Agriculture}. \textbf{188} pp. 106350 (2021), \url{https://www.sciencedirect.com/science/article/pii/S0168169921003677}
\bibitem{robo10}Alon, Y. \& Zhou, H. Multi-Agent Reinforcement Learning for Unmanned Aerial Vehicle Coordination by Multi-Critic Policy Gradient Optimization.  (2020)
\bibitem{robo11}Roghair, J., Niaraki, A., Ko, K. \& Jannesari, A. A Vision Based Deep Reinforcement Learning Algorithm for UAV Obstacle Avoidance. {\em Intelligent Systems And Applications}. pp. 115-128 (2022)
\bibitem{robo12}Nguyen, A., Pamuklu, T., Syed, A., Kennedy, W. \& Erol-Kantarci, M. Deep Reinforcement Learning for Task Offloading in UAV-Aided Smart Farm Networks. {\em 2022 IEEE Future Networks World Forum (FNWF)}. pp. 270-275 (2022)
\bibitem{robo13}Kurunathan, H., Li, K., Ni, W., Tovar, E. \& Dressler, F. Deep Reinforcement Learning for Persistent Cruise Control in UAV-aided Data Collection. {\em 2021 IEEE 46th Conference On Local Computer Networks (LCN)}. pp. 347-350 (2021)
\bibitem{robo14}Xu, B., Gao, F., Yu, C., Zhang, R., Wu, Y. \& Wang, Y. OmniDrones: An Efficient and Flexible Platform for Reinforcement Learning in Drone Control. {\em IEEE Robotics And Automation Letters}. pp. 1-7 (2024)
\bibitem{robo15}Tian, X., Pan, B., Bai, L., Wang, G. \& Mo, D. Fruit Picking Robot Arm Training Solution Based on Reinforcement Learning in Digital Twin. {\em Journal Of ICT Standardization}. \textbf{11}, 261-282 (2023)
\bibitem{robo16}Nethala, S. Autonomous Harvesting via Hierarchical Reinforcement Learning in Dynamic Environments. (Texas A\&M University-Corpus Christi,2023)
\bibitem{robo17}Yang, K., Zhang, Z., Cheng, H., Wu, H. \& Guo, Z. Domain centralization and cross-modal reinforcement learning for vision-based robotic manipulation. {\em International Journal Of Precision Agricultural Aviation}. \textbf{3} (2020)
\bibitem{robo18}Zeng, X., Zaenker, T. \& Bennewitz, M. Deep Reinforcement Learning for Next-Best-View Planning in Agricultural Applications. {\em 2022 International Conference On Robotics And Automation (ICRA)}. pp. 2323-2329 (2022)
\bibitem{robo19}Andriyanov, N. Development of Apple Detection System and Reinforcement Learning for Apple Manipulator. {\em Electronics}. \textbf{12} (2023), \url{https://www.mdpi.com/2079-9292/12/3/727}
\bibitem{robo20}Li, Y., Feng, Q., Zhang, Y., Peng, C., Ma, Y., Liu, C., Ru, M., Sun, J. \& Zhao, C. Peduncle collision-free grasping based on deep reinforcement learning for tomato harvesting robot. {\em Computers And Electronics In Agriculture}. \textbf{216} pp. 108488 (2024), \url{https://www.sciencedirect.com/science/article/pii/S0168169923008761}
\bibitem{robo21}Petrenko, V., Tebueva, F., Antonov, V. \& Gurchinsky, M. A Robotic Complex Control Method Based on Deep Reinforcement Learning of Recurrent Neural Networks for Automatic Harvesting of Greenhouse Crops. {\em Proceedings Of The 8th Scientific Conference On Information Technologies For Intelligent Decision Making Support (ITIDS 2020)}. pp. 340-346 (2020), \url{https://doi.org/10.2991/aisr.k.201029.064}
\bibitem{robo22}Martini, M., Cerrato, S., Salvetti, F., Angarano, S. \& Chiaberge, M. Position-Agnostic Autonomous Navigation in Vineyards with Deep Reinforcement Learning. {\em 2022 IEEE 18th International Conference On Automation Science And Engineering (CASE)}. pp. 477-484 (2022)
\bibitem{robo23}Yandun, F., Parhar, T., Silwal, A., Clifford, D., Yuan, Z., Levine, G., Yaroshenko, S. \& Kantor, G. Reaching Pruning Locations in a Vine Using a Deep Reinforcement Learning Policy. {\em 2021 IEEE International Conference On Robotics And Automation (ICRA)}. pp. 2400-2406 (2021)
\bibitem{robo24}Lin, G., Huang, P., Wang, M., Xu, Y., Zhang, R. \& Zhu, L. An Inverse Kinematics Solution for a Series-Parallel Hybrid Banana-Harvesting Robot Based on Deep Reinforcement Learning. {\em Agronomy}. \textbf{12} (2022), \url{https://www.mdpi.com/2073-4395/12/9/2157}
\bibitem{robo25}Wang, Y., He, Z., Cao, D., Ma, L., Li, K., Jia, L. \& Cui, Y. Coverage path planning for kiwifruit picking robots based on deep reinforcement learning. {\em Computers And Electronics In Agriculture}. \textbf{205} pp. 107593 (2023), \url{https://www.sciencedirect.com/science/article/pii/S0168169922009012}
\bibitem{robo26}Yang, J., Ni, J., Li, Y., Wen, J. \& Chen, D. The Intelligent Path Planning System of Agricultural Robot via Reinforcement Learning. {\em Sensors}. \textbf{22} (2022), \url{https://www.mdpi.com/1424-8220/22/12/4316}
\bibitem{robo27}Faryadi, S. \& Mohammadpour Velni, J. A reinforcement learning-based approach for modeling and coverage of an unknown field using a team of autonomous ground vehicles. {\em International Journal Of Intelligent Systems}. \textbf{36}, 1069-1084 (2021), \url{https://onlinelibrary.wiley.com/doi/abs/10.1002/int.22331}
\bibitem{robo28}Josef, S. \& Degani, A. Deep Reinforcement Learning for Safe Local Planning of a Ground Vehicle in Unknown Rough Terrain. {\em IEEE Robotics And Automation Letters}. \textbf{5}, 6748-6755 (2020)
\bibitem{robo29}Wiberg, V., Wallin, E., Nordfjell, T. \& Servin, M. Control of Rough Terrain Vehicles Using Deep Reinforcement Learning. {\em IEEE Robotics And Automation Letters}. \textbf{7}, 390-397 (2022)
\bibitem{robo30}Din, A., Ismail, M., Shah, B., Babar, M., Ali, F. \& Baig, S. A deep reinforcement learning-based multi-agent area coverage control for smart agriculture. {\em Computers And Electrical Engineering}. \textbf{101} pp. 108089 (2022), \url{https://www.sciencedirect.com/science/article/pii/S0045790622003445}
\bibitem{croma1}Elavarasan, D. \& Vincent, P. Crop Yield Prediction Using Deep Reinforcement Learning Model for Sustainable Agrarian Applications. {\em IEEE Access}. \textbf{8} pp. 86886-86901 (2020)
\bibitem{croma2}Elavarasan, D. \& Vincent, P. A reinforced random forest model for enhanced crop yield prediction by integrating agrarian parameters. {\em Journal Of Ambient Intelligence And Humanized Computing}. \textbf{12} pp. 10009-10022 (2021), Accessed on 2024-01-10
\bibitem{croma3}Wu, J., Tao, R., Zhao, P., Martin, N. \& Hovakimyan, N. Optimizing Nitrogen Management With Deep Reinforcement Learning and Crop Simulations. {\em Proceedings Of The IEEE/CVF Conference On Computer Vision And Pattern Recognition (CVPR) Workshops}. pp. 1712-1720 (2022,6)
\bibitem{croma4}Wang, C., Ye, Y., Tian, Y. \& Yu, Z. Classification of nutrient deficiency in rice based on CNN model with Reinforcement Learning augmentation. {\em 2021 International Symposium On Artificial Intelligence And Its Application On Media (ISAIAM)}. pp. 107-111 (2021)
\bibitem{croma5}Yassine, H., Roufaida, K., Shkodyrev, V., Abdelhak, M., Zarour, L. \& Khaled, R. Intelligent Farm Based on Deep Reinforcement Learning for optimal control. {\em 2022 International Symposium On INnovative Informatics Of Biskra (ISNIB)}. pp. 1-6 (2022)
\bibitem{croma6}Zheng, Z., Yan, P., Chen, Y., Cai, J. \& Zhu, F. Increasing Crop Yield Using Agriculture Sensing Data in Smart Plant Factory. {\em Security, Privacy, And Anonymity In Computation, Communication, And Storage}. pp. 345-356 (2021)
\bibitem{croma7}Sharma, G., Singh, A. \& Jain, S. DeepEvap: Deep reinforcement learning based ensemble approach for estimating reference evapotranspiration. {\em Applied Soft Computing}. \textbf{125} pp. 109113 (2022), \url{https://www.sciencedirect.com/science/article/pii/S156849462200388X}
\bibitem{croma8}Bouni, M., Hssina, B., Douzi, K. \& Douzi, S. Towards an Efficient Recommender Systems in Smart Agriculture: A deep reinforcement learning approach. {\em Procedia Computer Science}. \textbf{203} pp. 825-830 (2022), \url{https://www.sciencedirect.com/science/article/pii/S1877050922007293}, 17th International Conference on Future Networks and Communications / 19th International Conference on Mobile Systems and Pervasive Computing / 12th International Conference on Sustainable Energy Information Technology (FNC/MobiSPC/SEIT 2022), August 9-11, 2022, Niagara Falls, Ontario, Canada
\bibitem{croma9}Ashcraft, C. \& Karra, K. Machine Learning aided Crop Yield Optimization. {\em CoRR}. \textbf{abs/2111.00963} (2021), \url{https://arxiv.org/abs/2111.00963}
\bibitem{croma10}Madondo, M., Azmat, M., Dipietro, K., Horesh, R., Jacobs, M., Bawa, A., Srinivasan, R. \& O'Donncha, F. A SWAT-based Reinforcement Learning Framework for Crop Management.  (2023)
\bibitem{croma11}Overweg, H., Berghuijs, H. \& Athanasiadis, I. CropGym: a Reinforcement Learning Environment for Crop Management. {\em CoRR}. \textbf{abs/2104.04326} (2021), \url{https://arxiv.org/abs/2104.04326}
\bibitem{croma12}Kallenberg, M., Overweg, H., Bree, R. \& Athanasiadis, I. Nitrogen management with reinforcement learning and crop growth models. {\em Environmental Data Science}. \textbf{2} pp. e34 (2023)
\bibitem{croma13}Turchetta, M., Corinzia, L., Sussex, S., Burton, A., Herrera, J., Athanasiadis, I., Buhmann, J. \& Krause, A. Learning Long-Term Crop Management Strategies with CyclesGym. {\em Advances In Neural Information Processing Systems}. \textbf{35} pp. 11396-11409 (2022), \url{https://proceedings.neurips.cc/paper_files/paper/2022/file/4a22ceafe2dd6e0d32df1f7c0a69ab68-Paper-Datasets_and_Benchmarks.pdf}
\bibitem{water1}Tropea, M., Campoverde, L. \& De Rango, F. Exploiting Ai in Iot Smart Irrigation Management System: Reinforcement Learning vs Fuzzy Logic Models. {\em Available At SSRN 4149708}. (2022)
\bibitem{water2}Kelly, T., Foster, T. \& Schultz, D. Assessing the value of deep reinforcement learning for irrigation scheduling. {\em Smart Agricultural Technology}. \textbf{7} pp. 100403 (2024), \url{https://www.sciencedirect.com/science/article/pii/S277237552400008X}
\bibitem{water3}Ding, X. \& Du, W. DRLIC: Deep Reinforcement Learning for Irrigation Control. {\em 2022 21st ACM/IEEE International Conference On Information Processing In Sensor Networks (IPSN)}. pp. 41-53 (2022)
\bibitem{water4}Alibabaei, K., Gaspar, P., Assunção, E., Alirezazadeh, S. \& Lima, T. Irrigation optimization with a deep reinforcement learning model: Case study on a site in Portugal. {\em Agricultural Water Management}. \textbf{263} pp. 107480 (2022), \url{https://www.sciencedirect.com/science/article/pii/S0378377422000270}
\bibitem{water5}Alibabaei, K., Gaspar, P., Assunção, E., Alirezazadeh, S., Lima, T., Soares, V. \& Caldeira, J. Comparison of On-Policy Deep Reinforcement Learning A2C with Off-Policy DQN in Irrigation Optimization: A Case Study at a Site in Portugal. {\em Computers}. \textbf{11} (2022), \url{https://www.mdpi.com/2073-431X/11/7/104}
\bibitem{water6}Chen, M., Cui, Y., Wang, X., Xie, H., Liu, F., Luo, T., Zheng, S. \& Luo, Y. A reinforcement learning approach to irrigation decision-making for rice using weather forecasts. {\em Agricultural Water Management}. \textbf{250} pp. 106838 (2021), \url{https://www.sciencedirect.com/science/article/pii/S0378377421001037}
\bibitem{water7}Tao, R. Optimizing crop management with reinforcement learning, imitation learning, and crop simulations. (University of Illinois at Urbana-Champaign,2022,11)
\bibitem{water8}Zhou, N. Intelligent Control of Agricultural Irrigation Based on Reinforcement Learning. {\em Journal Of Physics: Conference Series}. \textbf{1601}, 052031 (2020,8), \url{https://dx.doi.org/10.1088/1742-6596/1601/5/052031}
\bibitem{water9}Hung, F. \& Yang, Y. Assessing Adaptive Irrigation Impacts on Water Scarcity in Nonstationary Environments—A Multi-Agent Reinforcement Learning Approach. {\em Water Resources Research}. \textbf{57}, e2020WR029262 (2021), \url{https://agupubs.onlinelibrary.wiley.com/doi/abs/10.1029/2020WR029262, e2020WR029262 2020WR029262}
\bibitem{water10}Gautron, R., Padrón, E., Preux, P., Bigot, J., Maillard, O. \& Emukpere, D. gym-DSSAT: a crop model turned into a Reinforcement Learning environment.  (2022)
\bibitem{green1}Wang, L., He, X. \& Luo, D. Deep Reinforcement Learning for Greenhouse Climate Control. {\em 2020 IEEE International Conference On Knowledge Graph (ICKG)}. pp. 474-480 (2020)
\bibitem{green2}Zhang, W., Cao, X., Yao, Y., An, Z., Xiao, X. \& Luo, D. Robust Model-based Reinforcement Learning for Autonomous Greenhouse Control. {\em Proceedings Of The 13th Asian Conference On Machine Learning}. \textbf{157} pp. 1208-1223 (2021,11,17), \url{https://proceedings.mlr.press/v157/zhang21e.html}
\bibitem{green3}Li, W., Wang, X., Jin, B., Luo, D. \& Zha, H. Structured Cooperative Reinforcement Learning With Time-Varying Composite Action Space. {\em IEEE Transactions On Pattern Analysis And Machine Intelligence}. \textbf{44}, 8618-8634 (2022)
\bibitem{green4}Ajagekar, A., Decardi-Nelson, B. \& You, F. Energy management for demand response in networked greenhouses with multi-agent deep reinforcement learning. {\em Applied Energy}. \textbf{355} pp. 122349 (2024), \url{https://www.sciencedirect.com/science/article/pii/S0306261923017130}
\bibitem{green5}Ajagekar, A., Mattson, N. \& You, F. Energy-efficient AI-based Control of Semi-closed Greenhouses Leveraging Robust Optimization in Deep Reinforcement Learning. {\em Advances In Applied Energy}. \textbf{9} pp. 100119 (2023), \url{https://www.sciencedirect.com/science/article/pii/S2666792422000373}
\bibitem{green6}Chen, L., Xu, L. \& Wei, R. Energy-Saving Control Algorithm of Venlo Greenhouse Skylight and Wet Curtain Fan Based on Reinforcement Learning with Soft Action Mask. {\em Agriculture}. \textbf{13} (2023), \url{https://www.mdpi.com/2077-0472/13/1/141}
\bibitem{green7}Decardi-Nelson, B. \& You, F. Improving Resource Use Efficiency in Plant Factories Using Deep Reinforcement Learning for Sustainable Food Production. {\em Chemical Engineering Transactions}. \textbf{103} pp. 79-84 (2023), \url{https://www.cetjournal.it/index.php/cet/article/view/CET23103014}
\bibitem{green8}Doan, Y., Ho, M., Nguyen, H. \& Han, H. Optimization of Spirulina sp. cultivation using reinforcement learning with state prediction based on LSTM neural network. {\em Journal Of Applied Phycology}. \textbf{33} pp. 2733-2744 (2021), Accessed on 2024-01-10
\bibitem{green9}Uyeh, D., Bassey, B., Mallipeddi, R., Asem-Hiablie, S., Amaizu, M., Woo, S., Ha, Y. \& Park, T. A Reinforcement Learning Approach for Optimal Placement of Sensors in Protected Cultivation Systems. {\em IEEE Access}. \textbf{9} pp. 100781-100800 (2021)
\bibitem{green10}Lu, Y., Gong, M., Li, J. \& Ma, J. Optimizing Controlled Environmental Agriculture for Strawberry Cultivation Using RL-Informer Model. {\em Agronomy}. \textbf{13} (2023), \url{https://www.mdpi.com/2073-4395/13/8/2057}
\bibitem{o1}Hribar, J., Marinescu, A., Chiumento, A. \& Dasilva, L. Energy-Aware Deep Reinforcement Learning Scheduling for Sensors Correlated in Time and Space. {\em IEEE Internet Of Things Journal}. \textbf{9}, 6732-6744 (2022)
\bibitem{o2}Nguyen, A., Pamuklu, T., Syed, A., Kennedy, W. \& Erol-Kantarci, M. Reinforcement Learning-Based Deadline and Battery-Aware Offloading in Smart Farm IoT-UAV Networks. {\em ICC 2022 - IEEE International Conference On Communications}. pp. 189-194 (2022)
\bibitem{o3}Ali, M., Alsaeedi, A., Shah, S., Yafooz, W. \& Malik, A. Energy Efficient Data Dissemination for Large-Scale Smart Farming Using Reinforcement Learning. {\em Electronics}. \textbf{12} (2023), \url{https://www.mdpi.com/2079-9292/12/5/1248}
\bibitem{o4}Naresh, M., Das, V., Saxena, P. \& Gupta, M. Deep reinforcement learning based QoE-aware actor-learner architectures for video streaming in IoT environments. {\em Computing}. \textbf{104} pp. 1527-1550 (2022), Accessed on 2024-01-10
\bibitem{o5}Ahmadi, M., Taghavirashidizadeh, A., Javaheri, D., Masoumian, A., Jafarzadeh Ghoushchi, S. \& Pourasad, Y. DQRE-SCnet: A novel hybrid approach for selecting users in Federated Learning with Deep-Q-Reinforcement Learning based on Spectral Clustering. {\em Journal Of King Saud University - Computer And Information Sciences}. \textbf{34}, 7445-7458 (2022), \url{https://www.sciencedirect.com/science/article/pii/S1319157821002226}
\bibitem{o6}Chen, H., Chen, Z., Lin, F. \& Zhuang, P. Effective Management for Blockchain-Based Agri-Food Supply Chains Using Deep Reinforcement Learning. {\em IEEE Access}. \textbf{9} pp. 36008-36018 (2021)
\bibitem{o7}Herabad, M. \& Afshar, N. Fuzzy-based Deep Reinforcement Learning for Frost Forecasting in IoT Edge-enabled Agriculture. {\em 2022 8th Iranian Conference On Signal Processing And Intelligent Systems (ICSPIS)}. pp. 1-5 (2022)
\bibitem{o8}Moeinizade, S., Hu, G. \& Wang, L. A reinforcement Learning approach to resource allocation in genomic selection. {\em Intelligent Systems With Applications}. \textbf{14} pp. 200076 (2022), \url{https://www.sciencedirect.com/science/article/pii/S2667305322000175}


\end{thebibliography}

\end{document}